%% file: main.tex
\documentclass{article}

\usepackage{arxiv}

\usepackage[utf8]{inputenc} 
\usepackage[T1]{fontenc}    
\usepackage{hyperref}       
\usepackage{url}            
\usepackage{booktabs}       
\usepackage{amsfonts}       
\usepackage{nicefrac}       
\usepackage{microtype}      
\usepackage{lipsum}		
\usepackage{graphicx}
\usepackage{natbib}
\usepackage{doi}

\usepackage{amsmath}
\usepackage{amsfonts}
\usepackage{array}

\title{Benchmarking the Domain Gap: Model Selection Instability Under Domain Shift in Video Capsule Endoscopy}

\author{
{Dan Hanson} \\
Department of Computer Science\\
Biomedical Perception \& Intelligence Lab\\
University of South Dakota\\
Vermillion, South Dakota, USA\\
\texttt{dan.hanson@usd.edu} 
\And
\href{https://orcid.org/0000-0002-8078-6730}
{\includegraphics[scale=0.06]{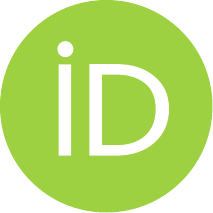}\hspace{1mm}Debesh Jha} \\
Department of Computer Science\\
Biomedical Perception \& Intelligence Lab\\
University of South Dakota\\
Vermillion, South Dakota, USA\\
\texttt{debesh.jha@usd.edu}
}

\renewcommand{\shorttitle}{Benchmarking the Domain Gap}

\hypersetup{
pdftitle={Benchmarking the Domain Gap: Model Selection Instability Under Domain Shift in Video Capsule Endoscopy},
pdfsubject={Computer Vision, Medical Imaging, Video Capsule Endoscopy},
pdfauthor={Dan Hanson, Debesh Jha},
pdfkeywords={video capsule endoscopy, domain shift, model selection, external validation},
}

\begin{document}
\maketitle

\begin{abstract}
Video capsule endoscopy (VCE) classification is typically evaluated within a
single dataset, yet clinical deployment demands robustness across acquisition
sources, labeling policies, and patient populations. We examine this gap using Kvasir-Capsule, Capsule Vision 2024 (CV2024), and a shared-label subset of Galar. We fine-tune a suite of general-domain pretrained backbones on the official Kvasir-Capsule folds under a standardized protocol and evaluate the same checkpoints on two non-source targets within a documented shared-label decision space. We find that the predictive value of in-domain ranking is target-dependent: Kvasir-Capsule ranking aligns more closely with Galar than with CV2024, while the two non-source targets agree only weakly. Consequently, the strongest in-domain backbone leads on one target yet falls to mid-pack on the
other, and no single evaluation target reliably predicts the others. A second
CV2024-trained configuration set reproduces this target-dependent instability.
We conclude that capsule endoscopy model selection should report cross-target
ranking stability rather than peak single-dataset performance.
\end{abstract}

\section{Introduction}
\label{sec:intro}

Video capsule endoscopy (VCE) enables non-invasive examination of the gastrointestinal tract and produces long image sequences that place substantial demands on clinical review. Automated frame classification has therefore become an active computer-vision problem, supported by public datasets such as Kvasir-Capsule~\cite{smedsrud2021kvasir}, the Capsule Vision 2024 (CV2024) challenge~\cite{handa2024capsulevision}, and Galar~\cite{lefloch2025galar}. Most published VCE classifiers, however, are developed and selected using performance within a single dataset. This creates an important model-selection question: does the model that ranks highest under source-domain validation remain the strongest model when the evaluation source changes?

This distinction matters because VCE datasets differ in more than class prevalence. They span different capsule systems, clinical sites, image characteristics, annotation policies, and label spaces. A backbone may therefore achieve strong internal performance by adapting effectively to one dataset, or combination of datasets, while ranking differently when evaluated on another. More generally, work on distribution shift and underspecification has shown that predictors with similar in-domain performance can behave differently outside the conditions under which they were selected~\cite{koh2021wilds,gulrajani2020domainbed,damour2022underspecification}. In medical imaging, acquisition-specific biases can further inflate internal performance when source-associated signals are shared between training and evaluation data~\cite{ongly2024hiddenbias}.

We study this problem in the setting of supervised fine-tuning of general-domain pretrained vision backbones. First, we benchmark 11 backbones under a standardized Kvasir-Capsule training configuration and evaluate them using the official video-isolated folds. The same trained checkpoints are then evaluated on compatible shared-label subsets of the CV2024 held-out test set and a subset of Galar. This design isolates backbone selection while holding the primary training configuration fixed. We then repeat the cross-target comparison from a second training source using a focused set of CV2024-trained configurations. Finally, we use multi-seed replication of representative Swin and CAFormer models to estimate whether observed cross-target changes exceed ordinary run-to-run variability.

A central difficulty in cross-dataset VCE evaluation is that the available datasets do not use identical label spaces. We therefore use a two-tier evaluation protocol. Cross-dataset comparisons are restricted to the intersection of eight compatible classes, with source predictions projected into a documented shared label space before scoring. Dataset-native or broader evaluable-class results are reported separately as capability context and are not used for cross-source ranking.

Our results show that the predictive value of in-domain ranking is target-dependent. Kvasir-Capsule backbone ranking agrees substantially with Galar but weakly with CV2024, while the two non-source targets also show weak
rank agreement. Thus, peak single-dataset performance and transferable model selection answer different questions: a model selected on one target cannot be assumed to retain its relative position under another.

Our contributions are as follows:
\begin{itemize}
    \item We provide a standardized official-fold benchmark of 11 general-domain pretrained vision backbones on Kvasir-Capsule, establishing a strong reference baseline for supervised fine-tuning.
    \item We evaluate the same Kvasir-trained checkpoints on CV2024 and Galar under a documented shared-label protocol and quantify backbone rank instability across three evaluation targets.
    \item We repeat the in-domain-to-external ranking analysis using selected CV2024-trained configurations, providing a second-source test of the observed model-selection instability.
\end{itemize}

\section{Related Work}
\label{sec:related-work}

\paragraph{Video Capsule Endoscopy Datasets.}
Public VCE datasets have enabled progress in automated abnormality classification, but they differ substantially in acquisition source, label space, annotation protocol, and split design. Kvasir-Capsule contains 117 videos, more than 4.7 million extractable frames, and 47,238 medically verified labeled frames across 14 classes~\cite{smedsrud2021kvasir}. Capsule Vision 2024 defines a 10-class abnormality-classification task using mixed-source training and validation data together with a held-out AIIMS test set~\cite{handa2024capsulevision, handa2024cv2024traindata, handa2024cv2024testdata}. Galar further expands the public VCE landscape with a multi-label, multicenter, multisystem dataset containing more than 3.5 million annotated frames across 29 labels~\cite{lefloch2025galar}. These datasets are complementary but not interchangeable: Kvasir-Capsule provides video-isolated official folds from a comparatively homogeneous acquisition source, CV2024 combines multiple source datasets in a harmonized challenge label space, and Galar provides a broad multisystem external source.

\paragraph{Domain Shift and External Validation.}
Distribution shift is a central obstacle for deployed machine-learning systems. WILDS formalized in-the-wild distribution-shift evaluation across domains including hospitals, geography, camera traps, and time, demonstrating substantial differences between in-distribution and out-of-distribution performance~\cite{koh2021wilds}. DomainBed further showed that inconsistent datasets, architectures, and model-selection criteria can obscure whether domain-generalization methods improve over strong empirical risk minimization (ERM) baselines~\cite{gulrajani2020domainbed}. These concerns are particularly relevant in medical imaging, where acquisition hardware, clinical site, protocol, preprocessing, patient population, and annotation practice can alter the test distribution. In VCE, analogous differences may arise from capsule system, illumination, image processing, bowel preparation, lesion prevalence, and clinical labeling conventions.

\paragraph{Shared-Class Cross-Dataset Evaluation.}
Cross-dataset evaluation can be complicated by non-identical native label
spaces, requiring the predictive task to be aligned before distributional
differences are compared. Restricted-class evaluation has precedent in
computer-vision robustness benchmarks: Hendrycks et al.\ evaluated an
ImageNet-1K classifier using only the outputs corresponding to a 200-class
target subset and reported similar clean-subset accuracy to a classifier
fine-tuned on those 200 classes~\cite{hendrycks2021natural}. In medical
imaging, Cohen et al.\ identified and canonically mapped 18 common labels
across chest X-ray datasets before measuring cross-domain
generalization~\cite{cohen2020crossdomain}. We follow the same general
principle of separating label-space harmonization from cross-dataset
evaluation. Our shared-class restriction is an evaluation operation rather
than a domain-adaptation or retraining procedure.

\paragraph{Shortcut Learning and Label Ambiguity.}
Medical imaging systems may exploit acquisition-specific or dataset-specific signals rather than clinically relevant features. Hidden acquisition biases can inflate apparent performance when training and evaluation data share non-causal source characteristics~\cite{ongly2024hiddenbias}. VCE introduces an additional challenge because fine-grained lesion categories can exhibit substantial visual overlap. Distinctions among erosion, erythema, ulcer, bleeding, and angioectasia may depend on lesion severity, frame quality, reader convention, and contextual information from neighboring video frames. Consequently, cross-dataset performance differences may reflect acquisition shift, label-ontology differences, or both. We therefore align only classes judged semantically compatible across datasets
and evaluate all direct cross-target comparisons in that shared task.

\paragraph{Underspecification, Seeds, and Model Selection.}
D'Amour et al. describe underspecification as a failure mode in which an ML pipeline can produce multiple predictors with similar held-out performance but substantially different behavior in deployment domains~\cite{damour2022underspecification}. This is directly relevant to VCE model selection: checkpoints, seeds, or architectures that appear comparable under internal validation may rank differently on another acquisition source. Arpit et al. similarly analyze instability across independently trained models and training trajectories in domain generalization, using rank correlation between in-domain validation and out-of-domain performance to study model-selection reliability~\cite{arpit2022ensemble}. We adopt this perspective by comparing backbone rankings across folds and external datasets and by using seed replication to estimate the scale of ordinary training variance.

\paragraph{Training Strategies Under Shift.}
Augmentation, reweighting, alternative objectives, and staged fine-tuning are
commonly used to improve robustness, yet domain-generalization benchmarks
caution that such interventions are not uniformly beneficial
\cite{gulrajani2020domainbed}. We therefore fix the training protocol in the
primary backbone comparison and treat the smaller CV2024 configuration set as
a second-source model-selection study rather than a factorial ablation.

\section{Methods}
\label{sec:methods}

\subsection{Evaluation Regimes}
\label{sec:evaluation-regimes}

\begin{figure}[t]
    \centering
    \includegraphics[width=\linewidth]{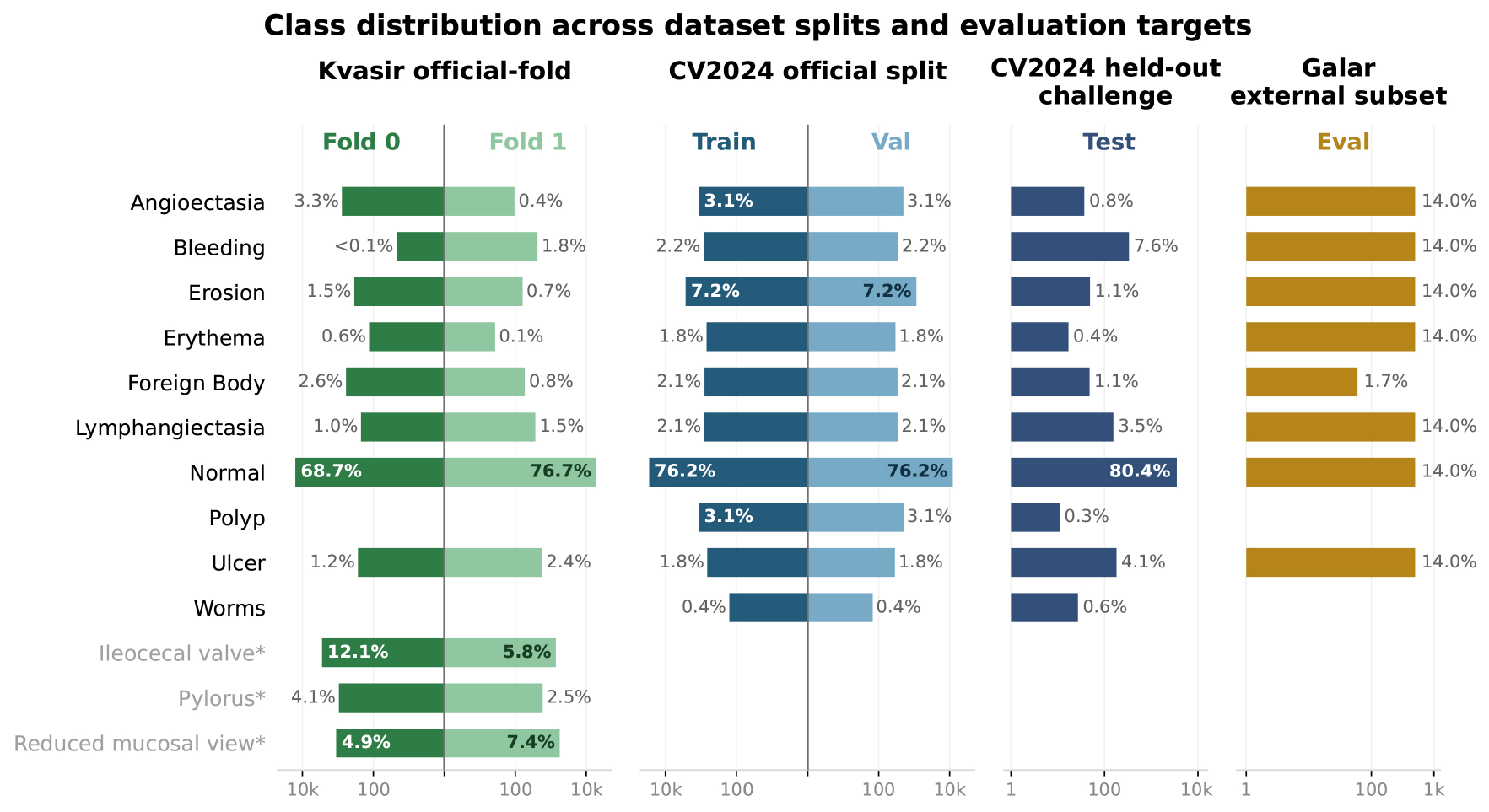}
    \caption{Class distributions across dataset splits and evaluation targets.
    Kvasir-Capsule shows the two official video-isolated folds; CV2024 shows
    the published train/validation split and held-out challenge test set.
    Galar shows the capped shared-8 evaluation subset, containing up to
    500 examples per class and all eligible Foreign Body examples.
    Kvasir-only classes are shown in gray and marked with an asterisk.
    Horizontal axes use logarithmic count scales; percentages indicate
    within-split class prevalence.}
    \label{fig:class-distributions}
\end{figure}

We study model ranking across three evaluation targets: Kvasir-Capsule
official-fold in-domain evaluation, the CV2024 held-out challenge test set, and
Galar external evaluation. These regimes differ in acquisition site, capsule
system, image characteristics, class prevalence, and native annotation
ontology (Table~\ref{tab:dataset-provenance} and
Fig.~\ref{fig:class-distributions}). CV2024 uses mixed-source training data, including limited AIIMS examples for Ulcer and Worms, followed by a separate private AIIMS held-out test set.
We therefore treat the three targets as distinct evaluation distributions
rather than interchangeable random splits.

\subsection{Evaluation Protocol}
\label{sec:evaluation-protocol}

The native label spaces of Kvasir-Capsule, CV2024, and Galar are not
identical. We therefore separate benchmark reporting from direct
cross-target comparison. The \textbf{native tier} reports performance in the
broader evaluable label space associated with each source dataset: the
11-class Kvasir-Capsule classification setting and the native 10-class
CV2024 challenge setting. These results preserve compatibility with
dataset-specific evaluation conventions but are not used for direct
cross-target ranking.

The \textbf{comparison tier} restricts evaluation to the eight semantically
aligned classes represented across all three datasets:
Angioectasia, Bleeding, Erosion, Erythema, Foreign Body,
Lymphangiectasia, Normal, and Ulcer. We refer to this common task as the
\emph{shared-8} label space. Cross-target rank correlations and rank-shift
analyses are computed only in this space.

Let $\mathcal{C}_s$ denote a model's source label space and
$\mathcal{C}$ the shared-8 comparison space. For input $x$, let
$p_i(x)$ denote the source softmax probability for
$i\in\mathcal{C}_s$. A fixed semantic mapping
$\phi:\mathcal{C}_s\rightarrow\mathcal{C}$ associates compatible source
classes with shared classes; source classes with zero training support or no
shared target are omitted. For $c\in\mathcal{C}$, mapped probability mass is

\begin{equation}
\tilde{p}_c(x)=
\sum_{\substack{i\in\mathcal{C}_s\\
\phi(i)=c,\;n_i>0}}
p_i(x),
\label{eq:shared-projection}
\end{equation}

where $n_i$ is source-class training support. We then normalize within the
comparison space,

\begin{equation}
q_c(x)=
\frac{\tilde{p}_c(x)}
{\sum_{u\in\mathcal{C}}\tilde{p}_u(x)}.
\label{eq:shared-renorm}
\end{equation}

Ground-truth examples outside $\mathcal{C}$ are excluded and predictions are
obtained as $\hat{y}=\arg\max_{c\in\mathcal{C}}q_c(x)$. Macro F1 and
multiclass Matthews correlation coefficient (MCC) are computed from the
restricted hard predictions; mean one-vs-rest AUROC (mAUROC) is computed from
the corresponding $q_c$ scores.

For one-to-one mappings, selecting and renormalizing the corresponding source
probabilities is exactly equivalent to masking non-shared logits before
softmax. The comparison tier therefore evaluates the classifier induced by
restricting a source-trained model to the known common task; no eight-class
retraining or target-domain parameter updates are performed. Related precedents
use restricted output subsets for robustness evaluation and canonical
shared-label alignment for medical cross-dataset evaluation
\cite{hendrycks2021natural,cohen2020crossdomain}.

For Kvasir-Capsule, Angiectasia maps to Angioectasia, Blood--fresh to
Bleeding, Foreign body to Foreign Body, and Normal clean mucosa to Normal;
Erosion, Erythema, Lymphangiectasia, and Ulcer map directly. After
zero-support classes are removed, these mappings are one-to-one for the
official Kvasir fold checkpoints. The mapping and class policy are fixed for
all models.

\subsection{Datasets and Label Mapping}
\label{sec:datasets-labels}

Table~\ref{tab:dataset-provenance} summarizes dataset provenance and acquisition
hardware. Kvasir-Capsule was acquired at B{\ae}rum Hospital using Olympus
Endocapsule 10~\cite{smedsrud2021kvasir}. CV2024 combines Kvasir-Capsule,
SEE-AI, KID, and private AIIMS data~\cite{handa2024capsulevision}; SEE-AI was
collected at Kyushu University Hospital using PillCam SB3
\cite{yokote2024seeai}, while the KID resource includes MiroCam imagery
\cite{koulaouzidis2017kid}. Galar spans two German clinical sites and multiple
Olympus and PillCam systems~\cite{lefloch2025galar}.

\input{tab_dataset_provenance}

For Galar comparison-tier evaluation, candidate frames are constructed
from the released per-frame labels using strict single-target logic. A frame
is assigned a shared-8 abnormality only when exactly one target label is
positive; frames with multiple target positives are excluded. Frames with no
target-positive label are assigned Normal only when located in the stomach,
small intestine, or colon and no other abnormality flag is active; otherwise
they are discarded. Eligible frames are sampled deterministically at the
frame level with seed 42, capped at 500 examples per class. Foreign Body has
60 eligible frames, yielding a 3,560-frame evaluation subset. Galar is used
only for external evaluation; no Galar training or validation split is
constructed.

All experiments use previously released VCE data described by the source
publications as anonymized or de-identified. Kvasir-Capsule was exempt from
regional ethics approval for anonymous export, Galar received ethics approval
BO-EK-534122022, and the AIIMS test data were collected under approval
IEC-666/05.08.2022
\cite{smedsrud2021kvasir,lefloch2025galar,handa2024capsulevision}.
No new patient data were collected for this study.

\subsection{Models and Training}
\label{sec:models-training}

The primary Kvasir-Capsule benchmark comprises 11 general-domain pretrained
vision backbones: ResNet-50 and ResNet-152 \cite{he2016resnet},
DenseNet-161 \cite{huang2017densenet}, EfficientNetV2-S
\cite{tan2021efficientnetv2}, ConvNeXt-Base \cite{liu2022convnext},
FocalNet-Base \cite{yang2022focalnet}, CAFormer-S18
\cite{yu2022metaformer}, MaxViT-Base \cite{tu2022maxvit}, ViT-B/16
\cite{dosovitskiy2020vit}, DINOv2 ViT-B/14 \cite{oquab2023dinov2},
and Swin-B \cite{liu2021swin}. Models are instantiated through
\texttt{timm} \cite{wightman2019timm} using general-domain pretrained
weights; no medical-domain pretraining or training from scratch is used.

All primary Kvasir baselines use the same minimal fine-tuning protocol: 25 training epochs, AdamW optimization with learning rate $1\times10^{-5}$ and weight decay $1\times10^{-4}$, cross-entropy loss, standard shuffled sampling, and no training augmentation. The complete network is therefore optimized end-to-end from the first epoch. Images are resized to the selected input resolution using Lanczos interpolation. Each architecture is evaluated at its selected operating resolution between 224 and 384 pixels; resolution is treated as part of the instantiated backbone configuration rather than as an isolated experimental variable.

A broader exploratory study evaluated staged fine-tuning, inverse-frequency weighted sampling, focal loss, and targeted photometric augmentation on the strongest backbone. Because these configurations were not exhaustively crossed across architecture, input resolution, and seed, we do not interpret them as a factorial ablation of individual training components. Selected configurations were instead used to construct a focused second study on CV2024.

For staged configurations, we use linear probing followed by full fine-tuning
(LP-FT)~\cite{kumar2022finetuning}. We freeze the pretrained backbone for
five epochs, then unfreeze the full network and apply a three-epoch linear
warmup followed by cosine decay. Weighted sampling uses inverse class-frequency sampling, and focal-loss configurations use $\gamma=2$~\cite{lin2017focal}. The geometric augmentation policy independently applies horizontal flipping, vertical flipping, and random rotation within $\pm15^\circ$, each with probability $p=0.5$. Additional photometric configurations apply either channel-wise gain and bias perturbation, whole-image color jitter, or a sparse mixed policy combining whole-image color jitter, channel-wise perturbation, and single-channel dropout.

\begin{figure}[t]
    \centering
    \includegraphics[width=\linewidth]{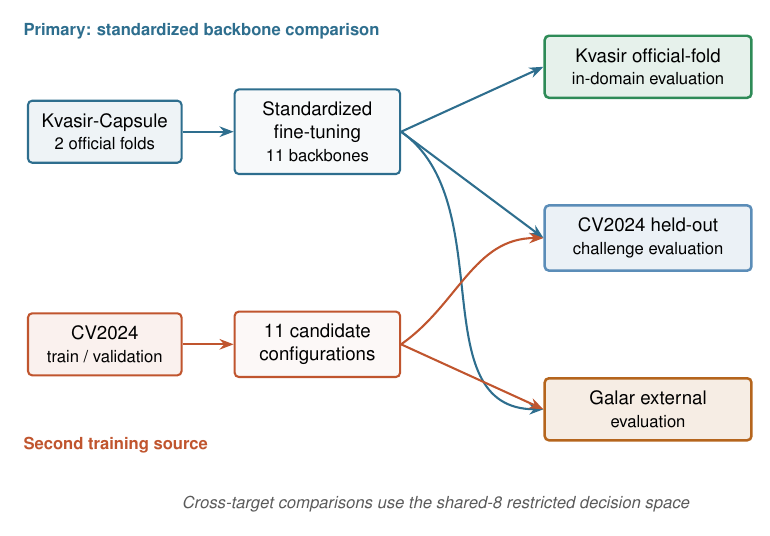}
    \caption{Overview of the experimental design. The primary study trains
    11 pretrained backbones under a standardized fine-tuning protocol on the
    two official Kvasir-Capsule folds and evaluates the resulting models on
    Kvasir, the held-out CV2024 challenge test set, and Galar. A second
    training source provides 11 CV2024-trained configurations
    evaluated on the held-out CV2024 test set and Galar. Direct cross-target
    comparisons use the shared-8 restricted decision space defined in
    Sec.~\ref{sec:evaluation-protocol}.}
    \label{fig:evaluation-design}
\end{figure}

Figure~\ref{fig:evaluation-design} summarizes the two training sources and
their evaluation paths. The primary standardized backbone comparison provides
the complete three-target analysis, while the focused CV2024-trained
set provides a second-source test of cross-target model-selection behavior.

\subsection{Evaluation Metrics}
\label{sec:metrics}

We report macro F1, multiclass Matthews correlation coefficient
(MCC)~\cite{matthews1975comparison,gorodkin2004kcategory}, and mean
one-vs-rest area under the receiver operating characteristic curve
(mAUROC)~\cite{fawcett2006roc}. Macro F1 gives equal weight to each
evaluated class, MCC summarizes multiclass confusion-matrix association,
and mAUROC measures class-wise ranking quality independently of the final
argmax decision.

Cross-target model-selection stability is evaluated using rank shifts and
Spearman rank correlation. Primary models are ranked by fold-averaged
shared-8 macro F1 on Kvasir and compared with their CV2024 and Galar
rankings. Fold0 versus Fold1 agreement provides a within-dataset reference,
while limited multi-seed Swin and CAFormer experiments provide additional
run-to-run context. Following prior work on model selection under
distribution shift
\cite{gulrajani2020domainbed,arpit2022ensemble,damour2022underspecification},
rank agreement is treated as distinct from absolute predictive performance.

\paragraph{Implementation and reproducibility.}
Models are implemented in PyTorch~\cite{paszke2019pytorch} and instantiated
with \texttt{timm}~\cite{wightman2019timm}. Metrics are computed with
\texttt{scikit-learn}~\cite{pedregosa2011scikit} and rank correlations with
\texttt{SciPy}~\cite{virtanen2020scipy}. Primary training uses
AdamW~\cite{loshchilov2019adamw} with learning rate $10^{-5}$, weight decay
$10^{-4}$, and 25 epochs; batch size is architecture-dependent because of
memory footprint. Python, NumPy, PyTorch, and CUDA seeds are fixed; cuDNN
deterministic mode is enabled, benchmarking and TF32 are disabled, and
deterministic algorithms are enforced. GPU autocast uses bfloat16; logits are
converted to float32 before softmax, and projection is performed in float64.
Exact checkpoint identifiers and per-configuration settings are provided in
the supplement.

\section{Results}
\label{sec:results}

\subsection{Kvasir-Capsule Official-Fold Benchmark}
\label{sec:kvasir-benchmark}

Before analyzing cross-target rank stability, we establish a
source-domain reference under the native 11-class Kvasir-Capsule
evaluation policy. Table~\ref{tab:kvasir-native11} reports
fold-averaged performance for the 11 standardized pretrained
backbones. All models use the common minimal fine-tuning protocol
described in Sec.~\ref{sec:models-training}; only the pretrained
backbone and its architecture-appropriate input resolution differ.

\begin{table}[t]
    \centering
    \small
    \setlength{\tabcolsep}{4.5pt}
    \caption{Kvasir-Capsule official-fold benchmark under the native
    11-class evaluable label policy. Values are averages across the
    two official fold directions. Best results are bold. Full per-fold
    results, shared-8 re-evaluations, and exact \texttt{timm}
    checkpoint identifiers are reported in the supplement.}
    \label{tab:kvasir-native11}
    \begin{tabular}{lrrrrr}
        \toprule
        Backbone & Params (M) & Input & Macro F1 & MCC & mAUROC \\
        \midrule
        Swin-B/384       & 86.9  & 384 & \textbf{0.3451} & \textbf{0.4970} & \textbf{0.8912} \\
        DINOv2 ViT-B/14  & 86.6  & 224 & 0.3089 & 0.4645 & 0.8760 \\
        CAFormer-S18/224 & 24.3  & 224 & 0.3076 & 0.4396 & 0.8679 \\
        ViT-B/16         & 85.8  & 224 & 0.2743 & 0.4091 & 0.8797 \\
        ConvNeXt-Base    & 87.6  & 224 & 0.2630 & 0.4328 & 0.8715 \\
        MaxViT-Base      & 118.7 & 224 & 0.2623 & 0.4155 & 0.8631 \\
        EfficientNetV2-S & 20.2  & 384 & 0.2440 & 0.3895 & 0.8263 \\
        FocalNet-Base    & 87.7  & 224 & 0.2430 & 0.4022 & 0.8459 \\
        DenseNet-161     & 26.5  & 224 & 0.2425 & 0.4293 & 0.7847 \\
        ResNet-50        & 23.5  & 224 & 0.1892 & 0.3599 & 0.7138 \\
        ResNet-152       & 58.2  & 224 & 0.1868 & 0.3173 & 0.7514 \\
        \bottomrule
    \end{tabular}
\end{table}

Swin-B/384 is the strongest standardized baseline under this protocol,
achieving a fold-averaged macro F1 of 0.3451, MCC of 0.4970, and
mAUROC of 0.8912. To the best of our knowledge, this is the strongest reported
Kvasir-Capsule result under a comparable official-fold evaluable-class protocol.
We treat this as a strong source-domain reference rather than evidence of external
robustness; the following analysis asks whether the models favored by
this benchmark remain favored when the evaluation target changes.

For cross-target analysis, the same checkpoints are re-evaluated in
the shared-8 comparison space. Native-11 and shared-8 Kvasir rankings
are highly consistent ($\rho=0.955$, $p<10^{-5}$), indicating that
shared-class restriction does not substantially reorder the
source-domain backbone benchmark.

\subsection{Backbone Rank Agreement Is Target-Dependent}
\label{sec:backbone-rank-instability}

The primary experiment fixes the Kvasir-Capsule training source and
fine-tuning protocol while varying the pretrained backbone. Models retain
their source-trained heads and are evaluated in shared-8 as defined in
Sec.~\ref{sec:evaluation-protocol}. Each backbone is ranked by the arithmetic
mean of its two official fold-direction macro-F1 scores, yielding one
architecture-level rank per target.

\begin{figure}[t]
    \centering
    \includegraphics[width=0.88\linewidth]{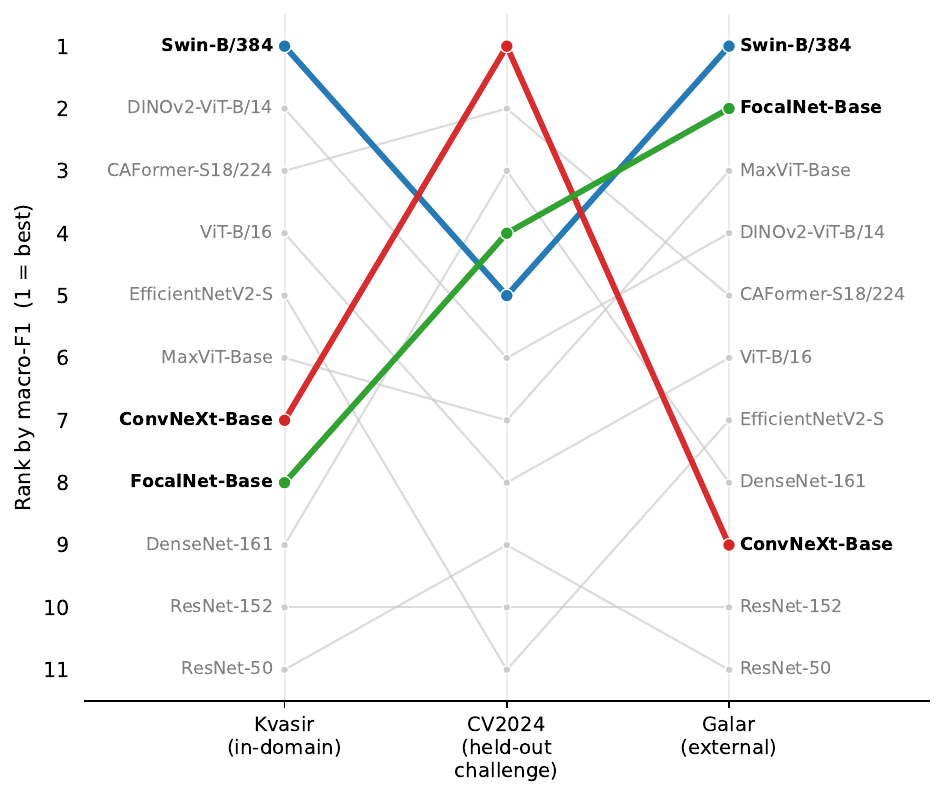}
    \caption{Backbone ranks by fold-averaged shared-8 macro F1 for
    Kvasir-trained models. Spearman agreement is $\rho=0.700$ for
    Kvasir--Galar, $\rho=0.191$ for Kvasir--CV2024, and $\rho=0.264$
    for CV2024--Galar. Highlighted trajectories are Swin-B/384
    ($1\!\rightarrow\!5\!\rightarrow\!1$), ConvNeXt-Base
    ($7\!\rightarrow\!1\!\rightarrow\!9$), and FocalNet-Base
    ($8\!\rightarrow\!4\!\rightarrow\!2$).}
    \label{fig:rank-shift-3target}
\end{figure}

Rank agreement varies markedly by target. Kvasir and Galar show stronger
agreement ($\rho=0.700$, $p=0.016$) than Kvasir and CV2024
($\rho=0.191$, $p=0.574$) or CV2024 and Galar
($\rho=0.264$, $p=0.433$). The highlighted trajectories show the practical
consequence: Swin-B/384 leads Kvasir and Galar but ranks fifth on CV2024;
ConvNeXt-Base leads CV2024 but ranks seventh and ninth on Kvasir and Galar;
and FocalNet-Base rises from eighth in-domain to second on Galar. Thus, even
when two targets show relatively strong aggregate rank agreement, a
source-domain position does not determine an individual backbone's rank on
another evaluation target.

\subsection{Second Training Source: CV2024-Trained Configurations}
\label{sec:cv2024-trained}

\begin{figure}[t]
    \centering
    \includegraphics[width=\linewidth]{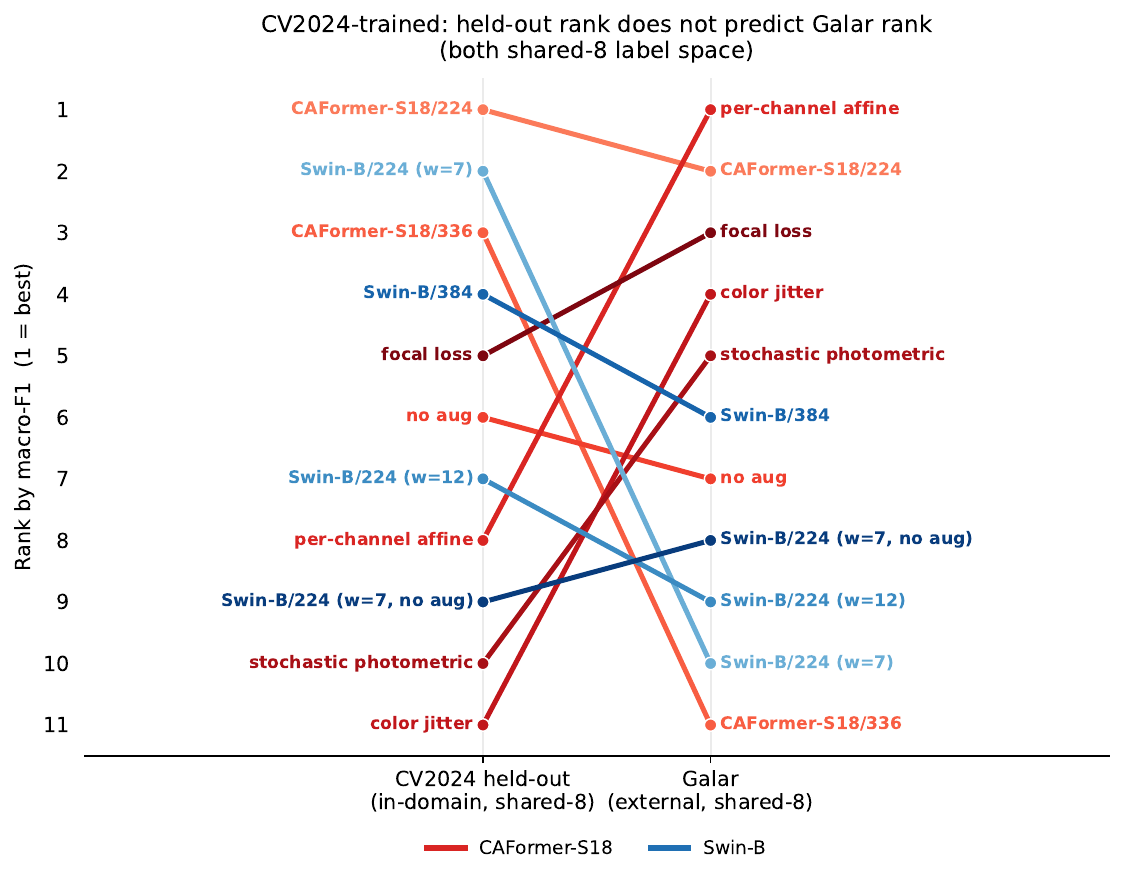}
    \caption{Macro-F1 rank changes for CV2024-trained configurations
    between the held-out CV2024 challenge evaluation and Galar external
    evaluation in the shared-8 label space.}
    \label{fig:cv2024-galar-ranks}
\end{figure}

We next evaluate 11 selected CAFormer and Swin training configurations motivated by exploratory work on class imbalance and photometric variability. Architecture and training factors are not fully crossed; this is therefore a
heterogeneous configuration set rather than a factorial ablation. Each configuration is trained on CV2024 and compared between the held-out CV2024 and Galar under shared-8 evaluation. Figure~\ref{fig:cv2024-galar-ranks} shows the macro-F1 rank changes, while Table~\ref{tab:cv2024-galar-shared8} reports the full metric panel.

\begin{table}[t]
\centering
\caption{Cross-target performance of CV2024-trained configurations in
the shared-8 comparison tier. Rows are ordered by held-out CV2024 macro F1.
F1 denotes macro F1 and mAUROC denotes mean one-vs-rest AUROC. All LP-FT
configurations use inverse-frequency weighted sampling; full fine-tuning uses
standard sampling and cross-entropy unless otherwise specified. Unlisted
augmentation indicates no training augmentation. Bold indicates the best value
for each metric and evaluation target. $w$ denotes the Swin attention-window
size.}
\label{tab:cv2024-galar-shared8}

\scriptsize
\setlength{\tabcolsep}{1.7pt}
\renewcommand{\arraystretch}{1.10}

\begin{tabular}{@{}
>{\raggedright\arraybackslash}p{0.18\linewidth}
>{\raggedright\arraybackslash}p{0.25\linewidth}
rrr@{\hspace{4pt}}rrr@{}}
\toprule
& &
\multicolumn{3}{c}{CV2024 held-out} &
\multicolumn{3}{c}{Galar external} \\
\cmidrule(lr){3-5}\cmidrule(lr){6-8}
Model & Training configuration &
F1 & MCC & mAUROC &
F1 & MCC & mAUROC \\
\midrule

CAFormer-S18/224 &
LP-FT; geometric augmentation &
\textbf{0.2840} & 0.1482 & 0.7319 &
0.2974 & 0.2203 & \textbf{0.7355} \\

Swin-B/224 ($w=7$) &
LP-FT; geometric augmentation &
0.2796 & \textbf{0.2522} & 0.7474 &
0.2776 & 0.2075 & 0.7325 \\

CAFormer-S18/336 &
LP-FT; geometric augmentation &
0.2662 & 0.1289 & 0.7456 &
0.2622 & 0.1884 & 0.7234 \\

Swin-B/384 ($w=12$) &
LP-FT; geometric augmentation &
0.2602 & 0.1926 & \textbf{0.7869} &
0.2924 & 0.2207 & 0.7282 \\

CAFormer-S18/224 &
Full fine-tuning; $\alpha$-weighted focal loss ($\gamma=2$) &
0.2561 & 0.1050 & 0.7343 &
0.2968 & 0.2451 & 0.7174 \\

CAFormer-S18/224 &
Full fine-tuning &
0.2271 & 0.0553 & 0.6677 &
0.2901 & 0.2533 & 0.6895 \\

Swin-B/224 ($w=12$) &
LP-FT; geometric augmentation &
0.2217 & 0.0939 & 0.7031 &
0.2872 & 0.2152 & 0.7307 \\

CAFormer-S18/224 &
Full fine-tuning; per-channel affine perturbation &
0.2061 & 0.0220 & 0.6816 &
\textbf{0.3131} & \textbf{0.2605} & 0.7012 \\

Swin-B/224 ($w=7$) &
LP-FT &
0.1901 & 0.0246 & 0.7453 &
0.2897 & 0.2310 & 0.7307 \\

CAFormer-S18/224 &
Full fine-tuning; stochastic photometric augmentation &
0.1877 & -0.0179 & 0.6522 &
0.2936 & 0.2417 & 0.7047 \\

CAFormer-S18/224 &
Full fine-tuning; image-level color jitter &
0.1795 & -0.0004 & 0.6420 &
0.2939 & 0.2330 & 0.6977 \\

\bottomrule
\end{tabular}
\end{table}

Across the 11 configurations, held-out CV2024 macro-F1 ranking showed weak agreement with
Galar (Spearman $\rho=-0.209$, $n=11$, $p=0.537$). The CV2024 macro-F1 leader,
CAFormer-S18/224 with LP-FT, weighted sampling, and geometric augmentation,
remained competitive on Galar but ranked second, whereas the per-channel affine
configuration rose from eighth on CV2024 to first on Galar. Conversely,
Swin-B/224 ($w=7$) fell from second to tenth by macro F1. Metric-specific model
selection also differed within the same target: the highest CV2024 MCC was
obtained by Swin-B/224 ($w=7$), while the highest CV2024 mAUROC was obtained
by Swin-B/384 ($w=12$). On Galar, the per-channel affine configuration achieved
both the highest macro F1 and MCC. These results do not identify a universally
superior training strategy; rather, they show that the selected model depends on
both the evaluation target and the model-selection metric.

\subsection{Within-Dataset and Run-to-Run Variability}
\label{sec:within-dataset-variability}

To contextualize cross-target reordering, we quantify fold- and seed-level
variability in shared-8. The two Kvasir folds show Spearman agreement of
$\rho=0.573$ ($p=0.07$, $n=11$), with mean macro F1 of $0.270$ on Fold0
and $0.197$ on Fold1. Thus, Kvasir--CV2024 agreement ($\rho=0.191$) is
descriptively lower than the observed within-Kvasir agreement, whereas
Kvasir--Galar agreement ($\rho=0.700$) is higher.

We additionally replicate representative Swin-B/384 and CAFormer-S18/224
configurations across three seeds and all three targets. Seed-level macro-F1
standard deviations range from $0.0060$ to $0.0121$ across family--target
pairs (mean $0.0092$). These variations are small relative to cross-target
score changes in the replicated families; mean Swin-B/384 macro F1, for
example, changes from $0.312$ on Kvasir to $0.190$ on CV2024. The seed
experiments therefore provide a magnitude reference for run-to-run variation,
not a suite-wide rank-stability guarantee.

\section{Discussion}
\label{sec:discussion}

The principal result is not that in-domain ranking never transfers. Rather,
its predictive validity is target-dependent. Kvasir backbone ordering was
closely aligned with Galar but weakly aligned with CV2024, despite applying the
same shared-class policy to all three evaluations. Moreover, CV2024 and Galar
showed weak agreement with each other. A single external validation target is
therefore not necessarily a universal proxy for transfer robustness: successful
selection against one shifted distribution may reveal little about another.

This distinction is important for interpreting model-selection studies under
distribution shift. An internally selected model may generalize well to a
particular target because the source and target preserve similar model
preferences, but this agreement is only observable after the target has been
evaluated. In the absence of prior knowledge that a deployment distribution
resembles Galar rather than CV2024, the Kvasir leaderboard alone provides no
basis for knowing which relationship will hold. We therefore argue for
reporting model-ranking stability across multiple independently sourced
evaluation targets rather than treating either in-domain performance or a
single external test set as sufficient evidence of transferable model
selection.

The FocalNet trajectory also illustrates that restricting external
evaluation to the highest-ranked source-domain models may overlook
competitive transfer candidates: FocalNet-Base ranked eighth on Kvasir
but second on Galar.

\section{Limitations}
\label{sec:limitations}

Shared-8 restriction evaluates source-trained classifiers under a common
decision space rather than reproducing models trained natively on eight
classes; supervision from source-only classes may influence learned
representations and decision boundaries. The restriction is fixed across
models, and native-11 and shared-8 Kvasir rankings agree closely
($\rho=0.955$), reducing concern that source-domain ordering is primarily a
label-space artifact. Nevertheless, our results characterize model selection
among the original source-trained classifiers, not independently trained
eight-class models.

Our design is asymmetric rather than fully crossed. The standardized backbone
suite is trained on Kvasir-Capsule, whereas the second training source contains
a focused CV2024 experimental set rather than the complete backbone suite. A fully
crossed two-source study would strengthen training-source conclusions but was
not feasible within our compute budget. The CV2024 configurations were also
selected from limited preliminary exploration rather than an exhaustive or
factorial search, so we do not assign causal effects or claim optimality for
individual training components.

Finally, three seed replication covers only two model families rather than the full suite,
and rank correlations are computed over only 11 backbones. Fold-level
variability therefore serves as the primary within-dataset reference, seed
results are interpreted as indicative, and individual correlation coefficients
should be interpreted cautiously.

\section{Conclusion}
\label{sec:conclusion}
We studied model-selection stability for video capsule endoscopy classification by training a standardized suite of general-domain backbones on Kvasir-Capsule and evaluating the same checkpoints across three targets in a documented shared-label decision space. The central finding is that the predictive value of an in-domain leaderboard is target-dependent: Kvasir ranking aligns well with Galar but poorly with CV2024, and the two external targets agree only weakly with each other, a pattern reproduced by a second CV2024-trained configuration set. No single evaluation target is a reliable proxy for the others, so a model selected against one distribution cannot be assumed to retain its standing under another. We therefore recommend that capsule endoscopy model selection report ranking stability across multiple independently sourced targets rather than peak single-dataset performance, and we release our shared-label evaluation protocol and standardized official-fold benchmark to support such reporting. Extending the standardized suite to a fully crossed multi-source training design is a clear direction for future work.

\bibliographystyle{plainnat}
\bibliography{main}
\end{document}

%% file: tab_dataset_provenance.tex
%

\begin{table}[t]
\centering
\scriptsize
\setlength{\tabcolsep}{1.8pt}
\renewcommand{\arraystretch}{1.08}
\caption{Acquisition provenance and native characteristics of the three dataset
regimes. Frame counts describe the released source data; the Galar experiments
use the 3,560-frame capped shared-8 subset described in
Sec.~\ref{sec:datasets-labels}. CV2024 counts are training/validation/test,
respectively. NR indicates not reported in the challenge paper.}
\label{tab:dataset-provenance}

\begin{tabular}{@{}
>{\raggedright\arraybackslash}p{0.115\linewidth}
>{\raggedright\arraybackslash}p{0.275\linewidth}
>{\raggedright\arraybackslash}p{0.225\linewidth}
>{\raggedright\arraybackslash}p{0.105\linewidth}
>{\raggedright\arraybackslash}p{0.155\linewidth}
r@{}}
\toprule
Dataset &
Acquisition source(s) &
Capsule system(s) &
Image size &
Frames &
Labels \\
\midrule

Kvasir-Capsule~\cite{smedsrud2021kvasir} &
B{\ae}rum Hospital, Vestre Viken Hospital Trust, Norway &
Olympus Endocapsule 10 (EC-S10; RE-10 recorder) &
$336{\times}336$ &
4,741,504 total; 47,238 labeled &
14 \\

CV2024~\cite{handa2024capsulevision, handa2024cv2024traindata, handa2024cv2024testdata} &
Kvasir, SEE-AI, KID, and private AIIMS; held-out test from AIIMS Delhi &
Mixed source systems; public source datasets include Endocapsule 10,
PillCam SB3, and MiroCam imagery; exact selected KID/AIIMS device mix NR &
$224{\times}224$ (challenge resize) &
58,124 (37,607 / 16,132 / 4,385) &
10 \\

Galar~\cite{lefloch2025galar} &
University Hospital Carl Gustav Carus, Dresden, and outpatient
gastroenterology practice, Dippoldiswalde, Germany &
Olympus Endocapsule 10; PillCam SB2, SB3, and Colon Capsule Endoscopy &
$336{\times}336$--$576{\times}576$ &
3,513,539 annotated; 3,560 used &
29 \\

\bottomrule
\end{tabular}
\end{table}

%% file: main.bbl
\begin{thebibliography}{34}
\providecommand{\natexlab}[1]{#1}
\providecommand{\url}[1]{\texttt{#1}}
\expandafter\ifx\csname urlstyle\endcsname\relax
  \providecommand{\doi}[1]{doi: #1}\else
  \providecommand{\doi}{doi: \begingroup \urlstyle{rm}\Url}\fi

\bibitem[Arpit et~al.(2022)Arpit, Wang, Zhou, and Xiong]{arpit2022ensemble}
Devansh Arpit, Huan Wang, Yingbo Zhou, and Caiming Xiong.
\newblock Ensemble of averages: Improving model selection and boosting performance in domain generalization.
\newblock In \emph{Advances in Neural Information Processing Systems}, volume~35, pages 8265--8277, 2022.
\newblock URL \url{https://proceedings.neurips.cc/paper_files/paper/2022/hash/372cb7805eaccb2b7eed641271a30eec-Abstract-Conference.html}.

\bibitem[Cohen et~al.(2020)Cohen, Hashir, Brooks, and Bertrand]{cohen2020crossdomain}
Joseph~Paul Cohen, Mohammad Hashir, Rupert Brooks, and Hadrien Bertrand.
\newblock On the limits of cross-domain generalization in automated x-ray prediction.
\newblock In \emph{Proceedings of the Third Conference on Medical Imaging with Deep Learning}, volume 121 of \emph{Proceedings of Machine Learning Research}, pages 136--155. PMLR, 2020.
\newblock URL \url{https://proceedings.mlr.press/v121/cohen20a.html}.

\bibitem[D'Amour et~al.(2022)D'Amour, Heller, Moldovan, Adlam, Alipanahi, Beutel, Chen, Deaton, Eisenstein, Hoffman, et~al.]{damour2022underspecification}
Alexander D'Amour, Katherine Heller, Dan Moldovan, Ben Adlam, Babak Alipanahi, Alex Beutel, Christina Chen, Jonathan Deaton, Jacob Eisenstein, Matthew~D. Hoffman, et~al.
\newblock Underspecification presents challenges for credibility in modern machine learning.
\newblock \emph{Journal of Machine Learning Research}, 23\penalty0 (226):\penalty0 1--61, 2022.
\newblock URL \url{https://www.jmlr.org/papers/v23/20-1335.html}.

\bibitem[Dosovitskiy et~al.(2021)Dosovitskiy, Beyer, Kolesnikov, Weissenborn, Zhai, Unterthiner, Dehghani, Minderer, Heigold, Gelly, Uszkoreit, and Houlsby]{dosovitskiy2020vit}
Alexey Dosovitskiy, Lucas Beyer, Alexander Kolesnikov, Dirk Weissenborn, Xiaohua Zhai, Thomas Unterthiner, Mostafa Dehghani, Matthias Minderer, Georg Heigold, Sylvain Gelly, Jakob Uszkoreit, and Neil Houlsby.
\newblock An image is worth 16x16 words: Transformers for image recognition at scale.
\newblock In \emph{International Conference on Learning Representations}, 2021.
\newblock URL \url{https://openreview.net/forum?id=YicbFdNTTy}.

\bibitem[Fawcett(2006)]{fawcett2006roc}
Tom Fawcett.
\newblock An introduction to {ROC} analysis.
\newblock \emph{Pattern Recognition Letters}, 27\penalty0 (8):\penalty0 861--874, 2006.
\newblock \doi{10.1016/j.patrec.2005.10.010}.

\bibitem[Gorodkin(2004)]{gorodkin2004kcategory}
Jan Gorodkin.
\newblock Comparing two {K}-category assignments by a {K}-category correlation coefficient.
\newblock \emph{Computational Biology and Chemistry}, 28\penalty0 (5--6):\penalty0 367--374, 2004.
\newblock \doi{10.1016/j.compbiolchem.2004.09.006}.

\bibitem[Gulrajani and Lopez-Paz(2021)]{gulrajani2020domainbed}
Ishaan Gulrajani and David Lopez-Paz.
\newblock In search of lost domain generalization.
\newblock In \emph{International Conference on Learning Representations (ICLR)}, 2021.
\newblock URL \url{https://openreview.net/forum?id=lQdXeXDoWtI}.

\bibitem[Handa et~al.(2024{\natexlab{a}})Handa, Mahbod, Schwarzhans, Woitek, Goel, Chhabra, Jha, Dhir, Sharma, Thakur, Gunjan, Kakarla, and Ramanathan]{handa2024cv2024testdata}
Palak Handa, Amirreza Mahbod, Florian Schwarzhans, Ramona Woitek, Nidhi Goel, Deepti Chhabra, Shreshtha Jha, Manas Dhir, Pallavi Sharma, Vijav Thakur, Deepak Gunjan, Jagadeesh Kakarla, and Balasubramanian Ramanathan.
\newblock Testing dataset of capsule vision 2024 challenge.
\newblock Figshare dataset, October 2024{\natexlab{a}}.
\newblock URL \url{https://doi.org/10.6084/m9.figshare.27200664.v3}.

\bibitem[Handa et~al.(2024{\natexlab{b}})Handa, Mahbod, Schwarzhans, Woitek, Goel, Chhabra, Jha, Sharma, Thakur, Dhir, Gunjan, Kakarla, and Raman]{handa2024cv2024traindata}
Palak Handa, Amirreza Mahbod, Florian Schwarzhans, Ramona Woitek, Nidhi Goel, Deepti Chhabra, Shreshtha Jha, Pallavi Sharma, Vijav Thakur, Manas Dhir, Deepak Gunjan, Jagadeesh Kakarla, and Balasubramanian Raman.
\newblock Training and validation dataset of capsule vision 2024 challenge.
\newblock Figshare dataset, July 2024{\natexlab{b}}.
\newblock URL \url{https://doi.org/10.6084/m9.figshare.26403469.v2}.

\bibitem[Handa et~al.(2024{\natexlab{c}})Handa, Mahbod, Schwarzhans, Woitek, Goel, Dhir, Chhabra, Jha, Sharma, Thakur, Chawla, Gunjan, Kakarla, and Raman]{handa2024capsulevision}
Palak Handa, Amirreza Mahbod, Florian Schwarzhans, Ramona Woitek, Nidhi Goel, Manas Dhir, Deepti Chhabra, Shreshtha Jha, Pallavi Sharma, Vijay Thakur, Simarpreet~Singh Chawla, Deepak Gunjan, Jagadeesh Kakarla, and Balasubramanian Raman.
\newblock Capsule vision 2024 challenge: Multi-class abnormality classification for video capsule endoscopy, 2024{\natexlab{c}}.
\newblock URL \url{https://arxiv.org/abs/2408.04940}.

\bibitem[He et~al.(2016)He, Zhang, Ren, and Sun]{he2016resnet}
Kaiming He, Xiangyu Zhang, Shaoqing Ren, and Jian Sun.
\newblock Deep residual learning for image recognition.
\newblock In \emph{Proceedings of the IEEE Conference on Computer Vision and Pattern Recognition}, pages 770--778, 2016.
\newblock URL \url{https://openaccess.thecvf.com/content_cvpr_2016/html/He_Deep_Residual_Learning_CVPR_2016_paper.html}.

\bibitem[Hendrycks et~al.(2021)Hendrycks, Zhao, Basart, Steinhardt, and Song]{hendrycks2021natural}
Dan Hendrycks, Kevin Zhao, Steven Basart, Jacob Steinhardt, and Dawn Song.
\newblock Natural adversarial examples.
\newblock In \emph{Proceedings of the IEEE/CVF Conference on Computer Vision and Pattern Recognition}, pages 15262--15271, 2021.
\newblock URL \url{https://openaccess.thecvf.com/content/CVPR2021/html/Hendrycks_Natural_Adversarial_Examples_CVPR_2021_paper.html}.

\bibitem[Huang et~al.(2017)Huang, Liu, van~der Maaten, and Weinberger]{huang2017densenet}
Gao Huang, Zhuang Liu, Laurens van~der Maaten, and Kilian~Q. Weinberger.
\newblock Densely connected convolutional networks.
\newblock In \emph{Proceedings of the IEEE Conference on Computer Vision and Pattern Recognition}, pages 4700--4708, 2017.
\newblock URL \url{https://openaccess.thecvf.com/content_cvpr_2017/html/Huang_Densely_Connected_Convolutional_CVPR_2017_paper.html}.

\bibitem[Koh et~al.(2021)Koh, Sagawa, Marklund, Xie, Zhang, Balsubramani, Hu, Yasunaga, Phillips, Gao, et~al.]{koh2021wilds}
Pang~Wei Koh, Shiori Sagawa, Henrik Marklund, Sang~Michael Xie, Marvin Zhang, Akshay Balsubramani, Weihua Hu, Michihiro Yasunaga, Richard~Lanas Phillips, Irena Gao, et~al.
\newblock {WILDS}: A benchmark of in-the-wild distribution shifts.
\newblock In \emph{Proceedings of the 38th International Conference on Machine Learning}, volume 139 of \emph{Proceedings of Machine Learning Research}, pages 5637--5664. PMLR, 2021.
\newblock URL \url{https://proceedings.mlr.press/v139/koh21a.html}.

\bibitem[Koulaouzidis et~al.(2017)Koulaouzidis, Iakovidis, Yung, Rondonotti, Kopylov, Plevris, Toth, Eliakim, Wurm~Johansson, Marlicz, Mavrogenis, Nemeth, Thorlacius, and Tontini]{koulaouzidis2017kid}
Anastasios Koulaouzidis, Dimitris~K. Iakovidis, Diana~E. Yung, Emanuele Rondonotti, Uri Kopylov, John~N. Plevris, Ervin Toth, Abraham Eliakim, Gabrielle Wurm~Johansson, Wojciech Marlicz, Georgios Mavrogenis, Artur Nemeth, Henrik Thorlacius, and Gian~Eugenio Tontini.
\newblock {KID Project}: An internet-based digital video atlas of capsule endoscopy for research purposes.
\newblock \emph{Endoscopy International Open}, 5\penalty0 (6):\penalty0 E477--E483, 2017.
\newblock \doi{10.1055/s-0043-105488}.

\bibitem[Kumar et~al.(2022)Kumar, Raghunathan, Jones, Ma, and Liang]{kumar2022finetuning}
Ananya Kumar, Aditi Raghunathan, Robbie Jones, Tengyu Ma, and Percy Liang.
\newblock Fine-tuning can distort pretrained features and underperform out-of-distribution.
\newblock In \emph{International Conference on Learning Representations}, 2022.
\newblock URL \url{https://openreview.net/forum?id=UYneFzXSJWh}.

\bibitem[Le~Floch et~al.(2025)Le~Floch, Wolf, McIntyre, Weinert, Palm, Volk, Herzog, Kirk, Steinh{\"a}user, Stopp, Geissler, Herzog, Sulk, Kather, Meining, Hann, Palm, Hampe, Herzog, and Brinkmann]{lefloch2025galar}
Maxime Le~Floch, Fabian Wolf, Lucian McIntyre, Christoph Weinert, Albrecht Palm, Konrad Volk, Paul Herzog, Sophie~Helene Kirk, Jonas~L. Steinh{\"a}user, Catrein Stopp, Mark~Enrik Geissler, Moritz Herzog, Stefan Sulk, Jakob~Nikolas Kather, Alexander Meining, Alexander Hann, Steffen Palm, Jochen Hampe, Nora Herzog, and Franz Brinkmann.
\newblock Galar -- a large multi-label video capsule endoscopy dataset.
\newblock \emph{Scientific Data}, 12\penalty0 (1):\penalty0 828, 2025.
\newblock \doi{10.1038/s41597-025-05112-7}.
\newblock URL \url{https://www.nature.com/articles/s41597-025-05112-7}.

\bibitem[Lin et~al.(2017)Lin, Goyal, Girshick, He, and Doll{\'a}r]{lin2017focal}
Tsung-Yi Lin, Priya Goyal, Ross Girshick, Kaiming He, and Piotr Doll{\'a}r.
\newblock Focal loss for dense object detection.
\newblock In \emph{Proceedings of the IEEE International Conference on Computer Vision}, pages 2980--2988, 2017.
\newblock URL \url{https://openaccess.thecvf.com/content_iccv_2017/html/Lin_Focal_Loss_for_ICCV_2017_paper.html}.

\bibitem[Liu et~al.(2021)Liu, Lin, Cao, Hu, Wei, Zhang, Lin, and Guo]{liu2021swin}
Ze~Liu, Yutong Lin, Yue Cao, Han Hu, Yixuan Wei, Zheng Zhang, Stephen Lin, and Baining Guo.
\newblock Swin transformer: Hierarchical vision transformer using shifted windows.
\newblock In \emph{Proceedings of the IEEE/CVF International Conference on Computer Vision}, pages 10012--10022, 2021.
\newblock URL \url{https://openaccess.thecvf.com/content/ICCV2021/html/Liu_Swin_Transformer_Hierarchical_Vision_Transformer_Using_Shifted_Windows_ICCV_2021_paper.html}.

\bibitem[Liu et~al.(2022)Liu, Mao, Wu, Feichtenhofer, Darrell, and Xie]{liu2022convnext}
Zhuang Liu, Hanzi Mao, Chao-Yuan Wu, Christoph Feichtenhofer, Trevor Darrell, and Saining Xie.
\newblock A {ConvNet} for the 2020s.
\newblock In \emph{Proceedings of the IEEE/CVF Conference on Computer Vision and Pattern Recognition}, pages 11976--11986, 2022.
\newblock URL \url{https://openaccess.thecvf.com/content/CVPR2022/html/Liu_A_ConvNet_for_the_2020s_CVPR_2022_paper.html}.

\bibitem[Loshchilov and Hutter(2019)]{loshchilov2019adamw}
Ilya Loshchilov and Frank Hutter.
\newblock Decoupled weight decay regularization.
\newblock In \emph{International Conference on Learning Representations}, 2019.
\newblock URL \url{https://openreview.net/forum?id=Bkg6RiCqY7}.

\bibitem[Matthews(1975)]{matthews1975comparison}
Brian~W. Matthews.
\newblock Comparison of the predicted and observed secondary structure of {T4} phage lysozyme.
\newblock \emph{Biochimica et Biophysica Acta (BBA) - Protein Structure}, 405\penalty0 (2):\penalty0 442--451, 1975.
\newblock \doi{10.1016/0005-2795(75)90109-9}.

\bibitem[Ong~Ly et~al.(2024)Ong~Ly, Unnikrishnan, Tadic, Patel, Duhamel, Kandel, Moayedi, Brudno, Hope, Ross, and McIntosh]{ongly2024hiddenbias}
Cathy Ong~Ly, Balagopal Unnikrishnan, Tony Tadic, Tirth Patel, Joe Duhamel, Sonja Kandel, Yasbanoo Moayedi, Michael Brudno, Andrew Hope, Heather Ross, and Chris McIntosh.
\newblock Shortcut learning in medical {AI} hinders generalization: Method for estimating {AI} model generalization without external data.
\newblock \emph{npj Digital Medicine}, 7\penalty0 (1):\penalty0 124, 2024.
\newblock \doi{10.1038/s41746-024-01118-4}.
\newblock URL \url{https://www.nature.com/articles/s41746-024-01118-4}.

\bibitem[Oquab et~al.(2024)Oquab, Darcet, Moutakanni, Vo, Szafraniec, Khalidov, Fernandez, Haziza, Massa, El-Nouby, et~al.]{oquab2023dinov2}
Maxime Oquab, Timoth{\'e}e Darcet, Th{\'e}o Moutakanni, Huy Vo, Marc Szafraniec, Vasil Khalidov, Pierre Fernandez, Daniel Haziza, Francisco Massa, Alaaeldin El-Nouby, et~al.
\newblock {DINOv2}: Learning robust visual features without supervision.
\newblock \emph{Transactions on Machine Learning Research}, 2024.
\newblock URL \url{https://openreview.net/forum?id=a68SUt6zFt}.

\bibitem[Paszke et~al.(2019)Paszke, Gross, Massa, Lerer, Bradbury, Chanan, Killeen, Lin, Gimelshein, Antiga, et~al.]{paszke2019pytorch}
Adam Paszke, Sam Gross, Francisco Massa, Adam Lerer, James Bradbury, Gregory Chanan, Trevor Killeen, Zeming Lin, Natalia Gimelshein, Luca Antiga, et~al.
\newblock {PyTorch}: An imperative style, high-performance deep learning library.
\newblock In \emph{Advances in Neural Information Processing Systems}, volume~32, pages 8024--8035, 2019.
\newblock URL \url{https://proceedings.neurips.cc/paper_files/paper/2019/hash/bdbca288fee7f92f2bfa9f7012727740-Abstract.html}.

\bibitem[Pedregosa et~al.(2011)Pedregosa, Varoquaux, Gramfort, Michel, Thirion, Grisel, Blondel, Prettenhofer, Weiss, Dubourg, et~al.]{pedregosa2011scikit}
Fabian Pedregosa, Ga{\"e}l Varoquaux, Alexandre Gramfort, Vincent Michel, Bertrand Thirion, Olivier Grisel, Mathieu Blondel, Peter Prettenhofer, Ron Weiss, Vincent Dubourg, et~al.
\newblock Scikit-learn: Machine learning in python.
\newblock \emph{Journal of Machine Learning Research}, 12:\penalty0 2825--2830, 2011.
\newblock URL \url{https://jmlr.org/papers/v12/pedregosa11a.html}.

\bibitem[Smedsrud et~al.(2021)Smedsrud, Thambawita, Hicks, Gjestang, Nedrejord, N{\ae}ss, Borgli, Jha, Berstad, Eskeland, Lux, Espeland, Petlund, Nguyen, Garcia-Ceja, Johansen, Schmidt, Toth, Hammer, de~Lange, Riegler, and Halvorsen]{smedsrud2021kvasir}
Pia~H. Smedsrud, Vajira Thambawita, Steven~A. Hicks, Henrik Gjestang, Oda~Olsen Nedrejord, Espen N{\ae}ss, Hanna Borgli, Debesh Jha, Tor Jan~Derek Berstad, Sigrun~L. Eskeland, Mathias Lux, H{\aa}vard Espeland, Andreas Petlund, Duc Tien~Dang Nguyen, Enrique Garcia-Ceja, Dag Johansen, Peter~T. Schmidt, Ervin Toth, Hugo~L. Hammer, Thomas de~Lange, Michael~A. Riegler, and P{\aa}l Halvorsen.
\newblock Kvasir-capsule, a video capsule endoscopy dataset.
\newblock \emph{Scientific Data}, 8\penalty0 (1):\penalty0 142, 2021.
\newblock \doi{10.1038/s41597-021-00920-z}.
\newblock URL \url{https://www.nature.com/articles/s41597-021-00920-z}.

\bibitem[Tan and Le(2021)]{tan2021efficientnetv2}
Mingxing Tan and Quoc~V. Le.
\newblock {EfficientNetV2}: Smaller models and faster training.
\newblock In \emph{Proceedings of the 38th International Conference on Machine Learning}, volume 139 of \emph{Proceedings of Machine Learning Research}, pages 10096--10106. PMLR, 2021.
\newblock URL \url{https://proceedings.mlr.press/v139/tan21a.html}.

\bibitem[Tu et~al.(2022)Tu, Talebi, Zhang, Yang, Milanfar, Bovik, and Li]{tu2022maxvit}
Zhengzhong Tu, Hossein Talebi, Han Zhang, Feng Yang, Peyman Milanfar, Alan Bovik, and Yinxiao Li.
\newblock {MaxViT}: Multi-axis vision transformer.
\newblock In \emph{Computer Vision--ECCV 2022}, pages 459--479. Springer, 2022.
\newblock URL \url{https://www.ecva.net/papers/eccv_2022/papers_ECCV/papers/136840453.pdf}.

\bibitem[Virtanen et~al.(2020)Virtanen, Gommers, Oliphant, Haberland, Reddy, Cournapeau, Burovski, Peterson, Weckesser, Bright, et~al.]{virtanen2020scipy}
Pauli Virtanen, Ralf Gommers, Travis~E. Oliphant, Matt Haberland, Tyler Reddy, David Cournapeau, Evgeni Burovski, Pearu Peterson, Warren Weckesser, Jonathan Bright, et~al.
\newblock {SciPy} 1.0: Fundamental algorithms for scientific computing in python.
\newblock \emph{Nature Methods}, 17:\penalty0 261--272, 2020.
\newblock \doi{10.1038/s41592-019-0686-2}.

\bibitem[Wightman(2019)]{wightman2019timm}
Ross Wightman.
\newblock Pytorch image models.
\newblock GitHub repository, 2019.
\newblock URL \url{https://github.com/huggingface/pytorch-image-models}.

\bibitem[Yang et~al.(2022)Yang, Li, Dai, and Gao]{yang2022focalnet}
Jianwei Yang, Chunyuan Li, Xiyang Dai, and Jianfeng Gao.
\newblock Focal modulation networks.
\newblock In \emph{Advances in Neural Information Processing Systems}, volume~35, 2022.
\newblock URL \url{https://proceedings.neurips.cc/paper_files/paper/2022/hash/1b08f585b0171b74d1401a5195e986f1-Abstract-Conference.html}.

\bibitem[Yokote et~al.(2024)Yokote, Umeno, Kawasaki, Fujioka, Fuyuno, Matsuno, Yoshida, Imazu, Miyazono, Moriyama, Kitazono, and Torisu]{yokote2024seeai}
Akihito Yokote, Junji Umeno, Keisuke Kawasaki, Shin Fujioka, Yuta Fuyuno, Yuichi Matsuno, Yuichiro Yoshida, Noriyuki Imazu, Satoshi Miyazono, Tomohiko Moriyama, Takanari Kitazono, and Takehiro Torisu.
\newblock Small bowel capsule endoscopy examination and open access database with artificial intelligence: The {SEE}-artificial intelligence project.
\newblock \emph{DEN Open}, 4\penalty0 (1):\penalty0 e258, 2024.
\newblock \doi{10.1002/deo2.258}.
\newblock URL \url{https://onlinelibrary.wiley.com/doi/10.1002/deo2.258}.

\bibitem[Yu et~al.(2024)Yu, Si, Zhou, Luo, Zhou, Feng, Yan, and Wang]{yu2022metaformer}
Weihao Yu, Chenyang Si, Pan Zhou, Mi~Luo, Yichen Zhou, Jiashi Feng, Shuicheng Yan, and Xinchao Wang.
\newblock {MetaFormer} baselines for vision.
\newblock \emph{IEEE Transactions on Pattern Analysis and Machine Intelligence}, 46\penalty0 (2):\penalty0 896--912, 2024.
\newblock \doi{10.1109/TPAMI.2023.3329173}.

\end{thebibliography}
